\documentclass[11pt,a4paper]{article}
\usepackage[hyperref]{emnlp2020}

\usepackage[utf8]{inputenc} 
\usepackage{microtype}
\usepackage{graphicx}
\usepackage{subfigure}
\usepackage{booktabs} 
\usepackage{comment}
\usepackage{hyperref}
\usepackage{mathtools}
\usepackage{color}
\usepackage{graphicx}
\usepackage{pifont}
\usepackage{CJKutf8}
\usepackage{url}
\usepackage{multicol,multirow}
\usepackage{makecell}
\usepackage{array}
\usepackage{bm}
\usepackage{parskip}
\usepackage[T1]{fontenc}    
\usepackage{booktabs}       
\usepackage{amsmath, amsthm}
\usepackage{amssymb}
\usepackage{amsfonts}

\usepackage{times}
\usepackage{float}
\usepackage{latexsym}

\usepackage{microtype}
\definecolor{ao}{rgb}{0.0, 0.5, 0.0}
\definecolor{asparagus}{rgb}{0.53, 0.66, 0.42}
\definecolor{amber}{rgb}{1.0, 0.49, 0.0}
\definecolor{alizarin}{rgb}{0.82, 0.1, 0.26}
\definecolor{applegreen}{rgb}{0.55, 0.71, 0.0}
\definecolor{amethyst}{rgb}{0.6, 0.4, 0.8}
\definecolor{auburn}{rgb}{0.43, 0.21, 0.1}
\aclfinalcopy 

\title{OpenViDial: A Large-Scale, Open-Domain Dialogue Dataset \\ with Visual Contexts}
\date{}

\author{
Yuxian Meng$^\clubsuit$, Shuhe Wang$^\spadesuit$, Qinghong Han$^\clubsuit$ \\ {\bf Xiaofei Sun$^\clubsuit$, Fei Wu$^\blacklozenge$,  Rongbin Ouyang$^\spadesuit$, Rui Yan$^\bigstar$ and Jiwei Li$^{\blacklozenge\clubsuit}$}\\
$^\blacklozenge$Zhejiang University,
  $^\spadesuit$Computer Center of Peking University \\
  $^\bigstar$Gaoling School of Artificial Intelligence, Renmin University of China \\ 
  $^\clubsuit$ Shannon.AI\\
  \{yuxian\_meng, qinghong\_han, xiaofei\_sun, jiwei\_li\}@shannonai.com\\
  wangshuhe@stu.pku.edu.cn, ouyang@pku.edu.cn, wufei@zju.edu.cn, ruiyan@ruc.edu.cn
}

\begin{document}
\maketitle

\begin{abstract}
When humans converse, what a speaker will say next significantly depends on what he sees. 
Unfortunately, existing dialogue  models 
generate dialogue utterances 
only based on 
preceding 
textual contexts, 
and  visual contexts are rarely considered.
This is due to 
 a lack of a 
 large-scale 
 multi-module 
 dialogue
  dataset with utterances paired with visual contexts.

In this paper, we release {\bf OpenViDial}, a large-scale  multi-module dialogue dataset.
The dialogue turns and visual contexts are extracted  from movies and TV series, 
where each dialogue turn is paired with the corresponding visual context in which it takes place. 
OpenViDial contains a total number of  1.1 million dialogue turns, and thus 1.1 million visual contexts stored in images.\footnote{Dataset is found at 
{\url{https://github.com/ShannonAI/OpenViDial}}.}  

\end{abstract}

\section{Introduction}

Giving machines the ability to converse like humans in the open domain is a key point towards passing the Turing test \citep{turing2009computing}, and developing 
open-domain dialogue agents is of growing interest \citep{,li2017adversarial,ghazvininejad2017knowledge,zhou2017emotional,gao2018neural,asghar2018affective,zhou2020design}. 
 Existing approaches towards developing open-domain dialogue agents are mostly data-driven, for which a large-scale dataset is first collected.
 The dataset usually consists of 
  millions of turns of dialogue utterances from real human 
 conversations. 
 A neural model is then trained on the dataset, learning to predict the upcoming dialogue turn conditioned on the previous textual contexts. 
\cite{li2016deep,li2016persona,zhang2018personalizing,huang2020challenges}
 
\begin{figure}
    \centering
    \includegraphics[scale=0.35]{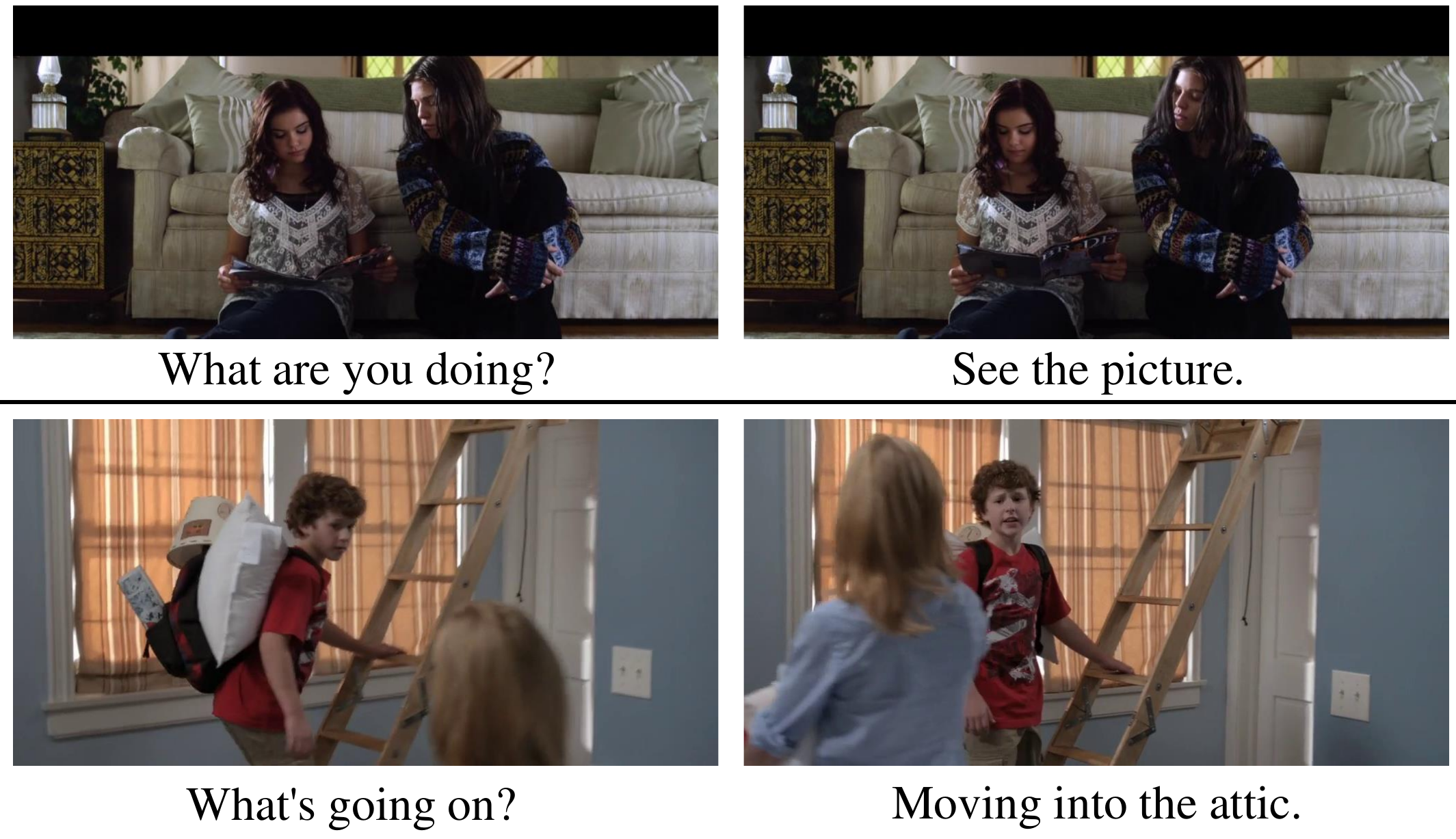}
    \caption{Two examples drawn from OpenViDial showing the necessity of considering visual contexts for dialogues.}
    \label{fig:intro_example}
\end{figure}

 One important aspect that existing open-domain dialogue models miss  is the consideration of multi-modal features in dialogue, especially visual features. 
When humans converse,  
what a speaker should say next significantly depends on what he sees. 
The granularity of visual features could be as large as 
the location that a conversation takes place in (e.g., a cafeteria or a theater), or as small as
 his dialogue partner's facial expressions. 
For example, in Figure \ref{fig:intro_example}, we present two short conversations where visual contexts are crucial. 
In both examples, if the model has no access to  visual information, 
it is hard to 
 correctly generate dialogue utterances  ``{\it see the picture}'' and ``{\it moving to the attic}'' in response to the preceding contexts.
 Unfortunately, existing dialogue  models 
generate dialogue utterances 
only based on 
preceding textual contexts and no visual contexts are considered. This is because of  
the lack of a 
 large-scale 
 multi-modal 
 dialogue dataset with utterances paired with visual context.

In this paper, 
we collect and release OpenViDial, a large-scale 
open-domain 
dialogue dataset with visual contexts.
The dialogue turns and visual contexts are extracted  from movies and TV series, 
where each dialogue turn is paired with the corresponding visual context in which it takes place. 
OpenViDial contains a total number of 1.1 million dialogue turns, and thus 1.1 million of visual contexts stored in images.

\section{Related Work}
\subsection{Existing Dialog Datasets}
\paragraph{Open Domain Dialog Datasets}
Over the past few years, 
various 
open-domain dialog datasets have been developed.
The OpenSubtitle dataset \citep{Tiedemann2009NewsFO,tiedemann-2012-parallel,Lison2016OpenSubtitles2016EL} consists of large-scale movie conversations extracted from the OpenSubtitle website. It includes a total number of 1,782 bitexts with 3.35G sentence fragments.
The Twitter Triple Corpus \citep{sordoni2015neural} consists of 4,232 Twitter conversation triples evaluated from 33K candidate triples by human raters, with 2,118 triples as tuning set and 2,114 as test set.
The Cornell Movie-Dialogs Corpus \citep{Danescu-Niculescu-Mizil+Lee:11a} contains a collection of fictional conversations extracted from raw movie scripts.
Other plain-text dialog datasets include the Ubuntu Dialog Corpus \citep{lowe2015ubuntu}, PersonaChat \citep{zhang2018personalizing}, EmpatheticDialogues \citep{rashkin2018towards}, etc.
The datasets described above only consist of texts in the form of dialogues, with no visual information included.

\paragraph{Visual Dialog Datasets}
The task of Visual Dialog is first introduced by \newcite{das2017visual}, where a model is required to answer a series of questions grounded in an image, given a dialog history and the image itself as contexts. 
Further, \newcite{das2017visual} released the VisDial v0.9 and v1.0 datasets as benchmarks. 
The v1.0 dataset 
contains 120K images from MS COCO\footnote{\url{http://mscoco.org/}} and each image is associated with 10 rounds of question-answer dialog, making up 1.2M examples in total.
The GuessWhat?! dataset \citep{devries2017guesswhat} 
focuses on high-level image understanding and is more goal-oriented: models need to locate an unknown object in an informative image scene by answering a sequence of ``yes or no'' questions. 
The CLEVER-Dialog \citep{kottur2019clevrdialog}  and MNIST-Dialog \citep{seo2017visual}  datasets are developed for diagnostic purposes. They are crafted to test the reasoning capability of visual dialog models based on the image and prior dialog turns. More recently, the Audio Visual Scene-Aware Dialog (AVSD) dataset \citep{hori2018endtoend,alamri2019audio} was introduced.  It contains more than 11,000 conversations paired with videos of human-centered activities, serving as a benchmark for the scene-aware video dialog task.
The datasets described above mainly focus on 
answering questions regarding an image or video, and thus are more concerned about question answering rather than dialogue generation.

\subsection{Dialogue Generation Models}
\paragraph{Open Domain Dialog Generation}
Building open-domain dialog systems that can converse with humans has a long history in natural language processing \citep{weizenbaum1966eliza,COLBY197537,Wallace2009}. Recent advances of neural networks have spurred great interests in developing neural-based data-driven dialog models \citep{vinyals2015neural,li2015diversity,dodge2016evaluating,serban2016hierarchical,zhao2017learning,xie2017data,lee2019convlab,ghandeharioun2019approximating,
li2020teaching,
han2020nonautoregressive,zhang2019dialogpt,roller2020recipes}. Built on top of sequence-to-sequence frameworks \citep{sutskever2014sequence,vaswani2017attention}, neural-based dialog models are able to generate coherent \citep{li2016deep,li2017adversarial,tian-etal-2017-make,bosselut-etal-2018-discourse,adiwardana2020towards}, diverse \citep{xu2018dp,baheti2018generating,ijcai2018-614}, personalized \citep{li2016persona,luan2017multi,zhang2018personalizing,zheng2019personalized,zheng2019pretraining,madotto2019personalizing}, informative \citep{shao2017generating,lewis-etal-2017-deal,ghazvininejad2017knowledge,young2017augmenting,zhao2019rethinking} and knowledge-fused \citep{10.1145/3340531.3411967,zhao-etal-2020-knowledge-grounded,he-etal-2020-amalgamating} responses, as well as bias toward different specific attributes or topics \citep{xing2016topic,zhou2017emotional,wang2017steering,niu2018polite,see2019makes}.

\paragraph{Visual Dialog Generation}
Since natural utterances and visual images are in different modalities, 
 attention mechanisms to model the interplay between conversational utterances and visual contents are widely used \citep{lu2017best,kottur2018visual,jiang2019dualvd,yang2019making,guo2019dual,niu2019recursive,kang2019dual,park2020multi,jiang2020dam}. 
\citet{seo2017visual} employed memories to store (attention, key) pairs that can be used to retrieve the most relevant attention maps for the current question in text.
\citet{schwartz2019factor} designed the factor graph attention model to connect an arbitrary number of modalities with attention flows. 
\citet{gan2019multi} proposed ReDAN, a recurrent dual attention network enhanced by a multi-step reasoning mechanism.
Techniques such as 
reinforcement learning \citep{das2017learning,wu2018you}, variational auto-encoders \citep{massiceti2018flipdial} and graph networks \citep{zheng2019reasoning,jiang2020kbgn}  have also been applied to deal with the visual dialog task.
Empowered 
by
 large-scale pretraining techniques, pretraining based models have made promising progress  \citep{lu2019vilbert,tan2019lxmert,su2019vl,alberti2019fusion,li2019unicodervl,li2019visualbert,chen2019uniter,wang2020vd,li2020oscar}, significantly boosting the performances in terms of different metrics.

\begin{table}
\centering
\small
\begin{tabular}{ll} \\\toprule
Number of turns &1.1M \\
Number of images & 1.1M\\
Vocab size before BPE & 70K \\
Vocab size after BPE & 30K \\
Average length of each episode & 14 \\
Average length of each turn & 7.6 \\\bottomrule
\end{tabular}
\caption{Detailed statistics for OpenViDial}
\label{tab:stats}
\end{table}

\section{Constructing OpenViDial}
In this section, we describe the details for OpenViDial construction. 
The main idea of 
dataset
 generation is to 
pair conversation scripts with  images in movies or TV series, and use these images as visual contexts for dialogue learning. 

We collect 
a raw dataset containing English
movies and TV series with a total length of roughly 8,000 hours. 
Each second of videos can be further divided into 20$\sim$40 frames, where each frame is an image.

\subsection{Subtitle Extraction based on OCR}
Because of the fact that only a small proportion of movies readily come with subtitle files, and that for most movies, subtitles are embedded in images, 
we need to build models to
extract conversation scripts from images. 
To build a conversation dataset with millions of turns of image-text pairs, it is prohibitively expensive and time-intensive  to employ human labors to separate each
image frame with embedded scripts. 
We thus rely on the technique of optical character recognition (OCR) for automatic extraction of conversation subtitles from movie images.\footnote{An alternative is to extract scripts from audios. We find extracting scripts using OCR from images obtains a much higher accuracy than 
speech recognition from audios. We  thus adopt the former strategy.  } 
We tailor the OCR model to the task of subtitle extraction, and achieves an almost perfect accuracy.

Existing open-sourced OCR models are not fit for our purpose since they are not tailored to subtitle extraction in the context of movies and TV series. We thus need to train our own OCR model.
\paragraph{Training Data Generation} We first synthesize the OCR training dataset, where we embed  texts into images to form training examples.  
To achieve this goal,
we first need to collect text-free images from raw videos, to which texts will be later added. 
This is done by running an existing open-sourced OCR model\footnote{\url{https://github.com/JaidedAI/EasyOCR}} on video images, and pick images with no text character identified by the model.
Since at this stage, our goal of identifying whether an image contains text character is a relatively easy task\footnote{This task can be make even easier by 
sacrificing recall (images without characters) for precision, by making sure that all selected images do not contain characters. 
},  a super accurate OCR model is not necessary and the open-sourced OCR model suffices to fulfill our need. 
With text-free images in hand, we pair them with texts. Texts are randomly selected from the CommonCrawl English corpus, then added to the images.
Texts  in images are generated using different fonts\footnote{\url{https://www.myfonts.com/WhatTheFont/}} and sizes. 
We generated a dataset containing about 10M images paired with texts.

\begin{table*}
\centering
\small
\scalebox{0.85}{
\begin{tabular}{lcccc} \\\toprule
{\bf Dataset} & {\bf Genre} & {\bf Multi-Modal?} & {\bf \# Sentences} &  {\bf \# Images}\\\midrule 
OpenSubtitles 2016 \citep{Lison2016OpenSubtitles2016EL} & Plain-text Dialog & \ding{55} &337M & -- \\
Cornell Movie-Dialogs \citep{Danescu-Niculescu-Mizil+Lee:11a} & Plain-text Dialog & \ding{55}  &0.3M &--  \\
VisDial v1.0 \citep{das2017visual}  & VQA & \checkmark & 2.4M & 120K  \\
Guess-What?! \citep{devries2017guesswhat}  & VQA & \checkmark & 0.8M & 66K \\
AVSD \citep{alamri2019audio} &  VQA & \checkmark & 152K & -- \\
OpenViDial (this work) & Visual+Text Dialog &  \checkmark & 1.1M & 1.1M  \\\bottomrule
\end{tabular}
}
\caption{A comparison of different datasets. VQA: Visual Question Answering.}
\label{tab:comparison}
\end{table*}

\paragraph{Model Training} 
Standard OCR training involves two stages, the {\it detection} of  the bounding box of texts, and the {\it recognition} of  characters. 
For detection, we use the PSE model as the backbone \cite{wang2019shape}, which is built upon 
 the FPN  model \cite{he2016identity} with ResNet  pre-trained on ImageNet dataset. 
 For recognition, we use the 
 Convolutional Recurrent Neural Network (CRNN)  model \cite{shi2016end}
as the backbone. 
We omit the details 
since the discussion on training OCR models is beyond the scope of this paper. 
We use a held-out dataset for evaluation, and the trained OCR model gets an accuracy higher than 99.98$\%$ at character level and 98.4$\%$ at the image/sentence level. 

\paragraph{Post Processing}
The trained OCR model is applied to  videos and TV series for script extraction. 
Since each second of the video  consists of 20$\sim$40 frames, most of which are 
nearly identical, 
we pick 3 frames for each second and discard the rest.  
We also construct an English vocabulary with top 200,000 words by frequency using a part of the CommonCrawl dataset, 
and remove images with unknown word from the vocabulary. This further helps us remove the influence from incorrect characters by the OCR model.
In addition, 
the following scenarios need to be handled: 
(1)
there are cases where a consecutive number of images are paired with the same texts. 
We only preserve the middle image and abandon the rest;
(2) There are cases where a full dialogue turn is truncated into 
 multiple consecutive
 images, with each image containing only part of the text in that
 dialogue 
 turn.
We train a simple discriminative model to identify whether a word in a context is the  end of a sentence. 
Using this model, we merge texts from multiple images into a single turn and pair the text with the middle image.

\subsection{Statistics for OpenViDial}
We collect a final dataset of 1.1M turns, where each turn consists of a sequence of words and an image.  
The size of the image is either $1280\times 720$ or 
$1920\times 1080$ based on different video resources. 
We employ the BPE tokenizer \citep{sennrich-etal-2016-neural} for text processing. 
 The detailed statistics  for OpenViDial
are shown in Table \ref{tab:stats}. 
We split the dataset into 1M/50K/50K for training, dev and test.

Table \ref{tab:comparison} shows the comparison between different datasets. 
Comparing against OpenSubtitles \citep{Lison2016OpenSubtitles2016EL}, OpenViDial has fewer sentences but contains
multi-modal features. 
Additionally, the OpenSubtitles dataset is an extremely noisy dataset, where 
consecutive lines
may not appear in the same conversation or scene,
and may not even be spoken by the same character. 
Comparing with other datasets with visual features, i.e., VisDial, Guess-What?! and AVSD, 
OpenViDial focuses more on dialogue learning rather than question answering.

\section{Conclusion}
In this paper, 
we  release OpenViDial, a large-scale 
open-domain 
dialogue dataset with visual contexts. 
In OpenViDial, 
each dialogue turn is paired with the corresponding visual context in which it takes place. 
  Our work marks an important step towards large-scale multi-modal dialogue learning. 

\bibliography{emnlp2020}
\bibliographystyle{acl_natbib}

\end{document}